\pdfoutput=1

\documentclass[11pt,table,dvipsnames]{article}

\usepackage[final]{acl}
\usepackage{times}
\usepackage{latexsym}

\usepackage[T1]{fontenc}

\usepackage[utf8]{inputenc}

\usepackage{microtype}

\usepackage{inconsolata}
\usepackage{microtype}
\usepackage{latexsym}
\usepackage{dsfont}
\usepackage{booktabs} 
\usepackage{CJKutf8}
\usepackage{graphicx}
\usepackage{bigstrut}
\usepackage{multirow}
\usepackage{amsmath}
\usepackage{multirow, makecell, caption}
\usepackage{floatrow}
\newfloatcommand{capbtabbox}{table}[][\FBwidth]
\usepackage{xcolor}
\usepackage[normalem]{ulem}
\usepackage{amssymb}
\usepackage{amsfonts}
\usepackage{multicol}
\usepackage{tipa}
\usepackage{amsmath}
\usepackage{bm}
\usepackage[ruled,linesnumbered]{algorithm2e}
\usepackage{tikz}
\usepackage{paralist}
\usepackage{IEEEtrantools}
\usepackage[switch]{lineno}
\usepackage{enumitem}
\usepackage{multirow, makecell, caption}
\usepackage{arydshln}
\usepackage{verbatim}
\usepackage[normalem]{ulem}
\usepackage{amssymb}
\usepackage{amsfonts}
\usepackage{multicol}
\usepackage{amssymb}
\usepackage{pifont}
\usepackage{csquotes}
\usepackage{xspace}
\usepackage{paralist}
\usepackage{mdwlist}
\usepackage{subcaption}
\usepackage{makecell}
\usepackage{array}
\usepackage{bm}
\usepackage{colortbl}
\usepackage{dsfont}
\usepackage{url}
\usepackage{booktabs} 
\usepackage{graphicx}
\usepackage{tabularx}
\usepackage{amsmath}
\usepackage{bbm}
\usepackage{pgfplots}
\usepackage{soul} 
\usepackage{listings}
\usepackage[tableposition=top]{caption}
\usepackage{color,xcolor} 

\newcommand{\ie}{\textit{i.e.}\xspace}
\newcommand{\eg}{\textit{e.g.}\xspace}

\newcommand{\method}{\texttt{SCUA}\xspace}

\makeatletter
\def\adl@drawiv#1#2#3{%
        \hskip.5\tabcolsep
        \xleaders#3{#2.5\@tempdimb #1{1}#2.5\@tempdimb}%
                #2\z@ plus1fil minus1fil\relax
        \hskip.5\tabcolsep}
\newcommand{\cdashlinelr}[1]{%
  \noalign{\vskip\aboverulesep
           \global\let\@dashdrawstore\adl@draw
           \global\let\adl@draw\adl@drawiv}
  \cdashline{#1}
  \noalign{\global\let\adl@draw\@dashdrawstore
           \vskip\belowrulesep}}
\makeatother

\title{\textit{Boosting Scientific Concepts Understanding}:\\ Can Analogy from Teacher Models Empower Student Models?}

\author{
 Siyu Yuan\textsuperscript{\rm $\heartsuit$}\thanks{~~Equal contribution.},
Cheng Jiayang\textsuperscript{\rm $\spadesuit$}\footnotemark[1],
Lin Qiu\textsuperscript{\rm $\diamondsuit$}\thanks{~~Corresponding authors.},
 \textbf{Deqing Yang}\textsuperscript{\rm $\heartsuit$}\footnotemark[2]\\
\textsuperscript{\rm $\heartsuit$}School of Data Science, Fudan University\\
\textsuperscript{\rm $\spadesuit$}The Hong Kong University of Science and Technology\\
\textsuperscript{\rm $\diamondsuit$}Shanghai Jiao Tong University\\
\texttt{syyuan21@m.fudan.edu.cn}, 
\texttt{jchengaj@cse.ust.hk}\\
\texttt{lqiu@apex.sjtu.edu.cn},
\texttt{deqingyang@fudan.edu.cn}\\
}

\begin{document}

\maketitle

\begin{abstract}

Analogical reasoning plays a critical role in human cognition, enabling us to understand new concepts by associating them with familiar ones.
Previous research in the AI community has mainly focused on identifying and generating analogies and then examining their quality under human evaluation, which overlooks the practical application of these analogies in real-world settings.
Inspired by the human education process, in this paper, we propose to investigate how analogies created by teacher language models (LMs) can assist student LMs in understanding scientific concepts, thereby aligning more closely with practical scenarios.
Our results suggest that free-form analogies can indeed aid LMs in understanding concepts.
Additionally, analogies generated by student LMs can improve their own performance on scientific question answering, demonstrating their capability to use analogies for self-learning new knowledge.
Resources are available at \url{https://github.com/siyuyuan/SCUA}.
\end{abstract}

\section{Introduction}
\label{sec:intro}
\begin{figure}[t]
    \centering
    \small
    \includegraphics[width=\linewidth]{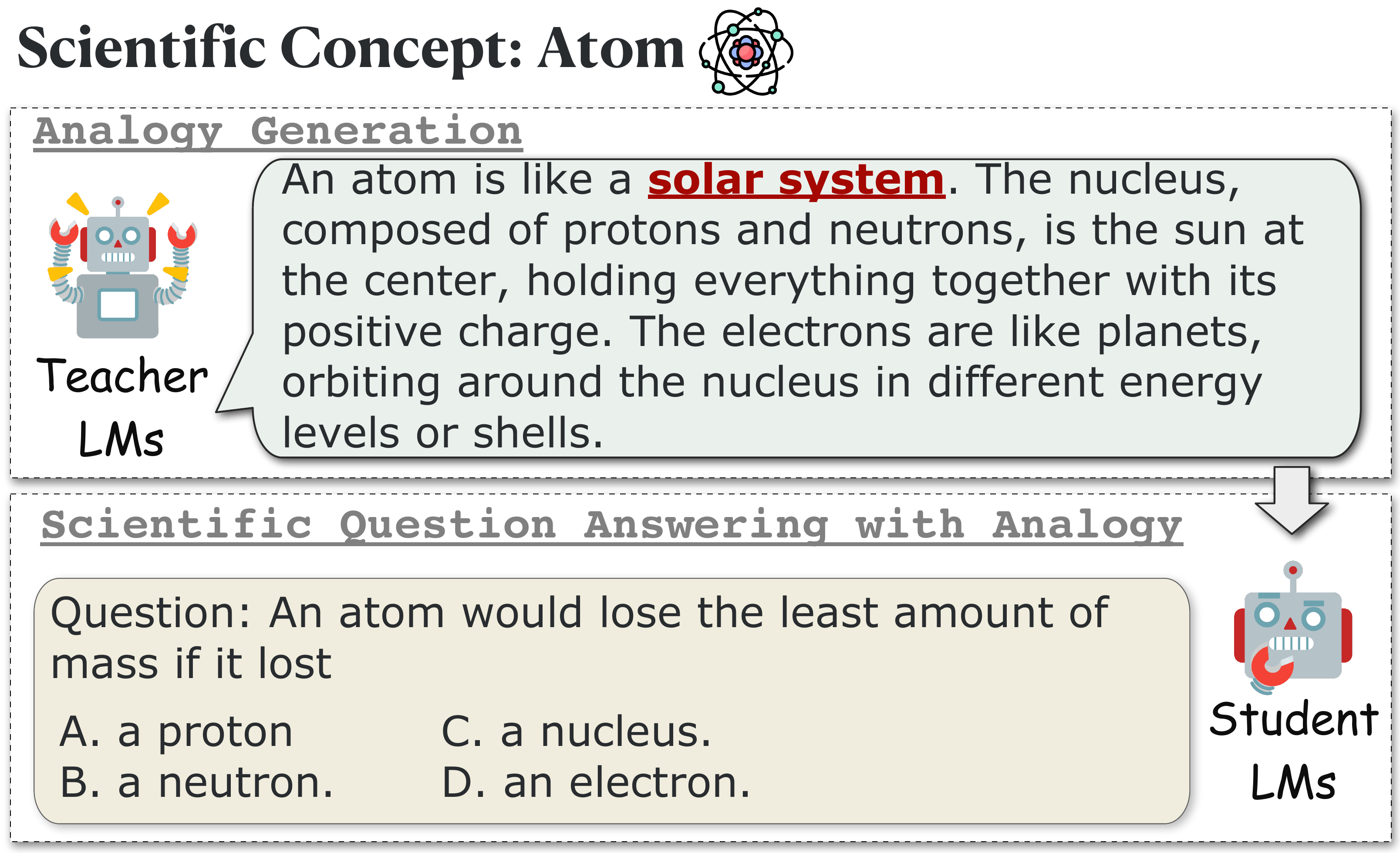}
    \caption{An example of the \method task. Given a scientific concept (\ie, Atom), we ask teacher LMs to generate an analogy to explain the concept and then let student LMs answer the related scientific questions around this concept, both with and without the aid of the generated analogy.}
    \label{fig:front}
\end{figure}

Analogy plays a crucial role in human cognition, facilitating the understanding of complex and unfamiliar concepts by relating them to familiar ones~\cite{bunge1981analogy,glynn1989analogical,hofstadter2001analogy,bartha2013analogy}. 
For example, Figure~\ref{fig:front} illustrates how using the solar system as an analogy can enhance understanding of the complex structure of atoms. 
Given its significant value across various fields, including creativity~\cite{10.1145/3530013} and education~\cite{richland2015analogy,thagard1992analogy}, the topic of analogy has been drawing significant research attention in the AI community.

Traditional research on analogy primarily focuses on evaluating~\cite{siword,schluter-2018-word,czinczoll-etal-2022-scientific,chen-etal-2022-e} and enhancing~\cite{ushio-etal-2021-bert,yuan2023analogykb} the analogical reasoning capabilities of language models (LMs) in word analogies (\eg, ``king is to man as queen is to woman'').
Recent advancements in large language models (LLMs)~\cite{openai2022chatgpt,openai2023gpt4} have shifted this focus from simple word analogies to exploring analogies between more complex situations such as systems~\cite{yuan-etal-2023-beneath}, processes~\cite{bhavya2022analogy,sultan-shahaf-2022-life,ding2023fluid,sultan2024parallelparc}, paragraphs~\cite{webb2022emergent,wijesiriwardene-etal-2023-analogical}, and stories~\cite{jiayang-etal-2023-storyanalogy}.
However, these studies mainly examine whether LLMs can generate appropriate analogies under human evaluation without thoroughly assessing the practical functionality of the generated analogies in real-world scenarios.

In this paper, drawing on principles of human education, we propose the \method, \ie, \textbf{S}cientific \textbf{C}oncept \textbf{U}nderstanding with \textbf{A}nalogy task, which aims to investigate whether analogies generated by teacher LMs can assist student LMs in understanding scientific concepts.
Specifically, as shown in Figure~\ref{fig:front}, given a scientific concept, we initially prompt teacher LMs, (\eg, GPT-4~\cite{openai2023gpt4} and Claude~\cite{anthropic2024claude3}), to generate an analogy that explains the concept. 
Then, we collect related scientific questions around this concept from the database and let student LMs (\eg, GPT-3.5~\cite{openai2022chatgpt} and Vicuna~\cite{chiang2023vicuna}) attempt to answer these questions, with and without the use of the generated analogy.

Under this setting, we conduct extensive experiments to evaluate strong and weak LMs with different analogy types.
The main findings are as follows:
\begin{itemize}[leftmargin=*]
    \item Analogies indeed help LMs understand scientific concepts, improving their ability to answer scientific questions. 
    \item Although word analogies generated by teacher LMs reveal the highest quality, the more sophisticated structured and free-text analogies bring higher improvements to the student LMs, suggesting that future work can focus on enhancing the quality of structured and free-text analogies.
    \item Analogy generated by student LMs can boost their own performance on scientific quizzes, illustrating their ability to leverage analogies for self-learning new knowledge.
\end{itemize}

\section{Related Work}
\label{sec:related}

\paragraph{Analogical Reasoning}
Analogical reasoning has long interested the AI community~\cite{davies1985analogy,gentner2011computational,mitchell2021abstraction}. 
Traditional research has focused on word analogies, examining linear relationships between words~\cite{gladkova-etal-2016-analogy,schluter-2018-word,fournier-etal-2020-analogies,ushio-etal-2021-bert}. 
With the development of LLMs~\cite{openai2022chatgpt,openai2023gpt4,geminiteam2023gemini}, there has been a shift toward exploring analogies between situations, establishing mappings between concepts across two domains based on shared relational structures~\cite{sultan-shahaf-2022-life,ding2023fluid,jiayang-etal-2023-storyanalogy,sultan2024parallelparc}.  
Compared to these studies, our research is the first to explore how analogies generated by teacher LMs can aid student LMs in understanding scientific concepts, which is more aligned with real-world scenarios.

\paragraph{Explanation Generation}
With the rising capabilities of LLMs, prior research has adopted methods, \eg, Chain of Thought (CoT)~\cite{wei2022chain,zhang2023automatic}, to generate a reasoning process before answering.
Due to the relatively limited capabilities of smaller LMs, some studies employ knowledge distillation, which involves generating reasoning samples using larger LMs to instruct smaller models~\cite{wang2023pinto,hsieh-etal-2023-distilling,wang-etal-2023-scott,lin-etal-2023-beneath,he2024complex}.
Compared to these studies, our work is the first to explore explanations with analogical reasoning in understanding scientific concepts.

\section{\method Task}

\label{sec:method}

\begin{table*}[t]
  \centering
  \small
    \begin{tabular}{ll}
    \toprule
    \multicolumn{2}{c}{\textbf{Scientific Concept}: Thermal Equilibrium} \\
    \midrule
    \textbf{Free-Form Analogy} & \makecell[l]{
    Imagine a group of children, each holding a different number of balloons and standing in a room. Over\\ time, they start trading balloons to balance out their amounts until each child is holding roughly the same \\number. Thermal equilibrium works similarly with temperature. If you place a hot object and a cold\\
    object close together, heat (like the balloons) will transfer from the hot object to the cold one until both...
    } \\
    \midrule
    \textbf{Structure Analogy} & \makecell[l]{1. Hot and cold objects correspond to weights on a scale.\\2. Heat transfer corresponds to weight redistribution. \\3. The point of equilibrium corresponds to the balance point on a scale.\\4. The cessation of heat flow corresponds to the stillness of the scale.} \\
    \midrule
    \textbf{Word Analogy} & Thermal Equilibrium can be analogous to a Balancing Scale \\
    \bottomrule
    \end{tabular}%
    \caption{Examples of three types of analogy for a scientific concept.}
  \label{tab:example}%
\end{table*}%

\subsection{Task Formulation}

As illustrated in Figure~\ref{fig:front}, given a scientific concept $C$, we initially ask teacher LMs to generate analogies $C_{A}$ to explain this concept. 
Then we will give a scientific question $Q$ and $m$ candidate answer $A = \{A_i\}_{i=1}^m$, which is related to the scientific concept $C$.
The ultimate goal of student LMs is to make the correct choice $\mathcal{Y}$ for $\mathcal{X} = (Q, A, C_{A})$.

\subsection{Analogy Generation from Teacher LMs}\label{sec:analogy_type}

\paragraph{Scientific Concept Extraction}
Current scientific question answering (QA) datasets rarely explicitly contain the related concepts for reference.
Thus, given a scientific question, we adopt GPT-4 to extract one scientific concept related to this question.
Next, we employ three annotators to evaluate and improve the quality of the concepts.
Then, teacher LMs generate analogies for these concepts.\footnote{The extraction prompt for GPT-4 is shown in Appendix~\ref{sec:extraction}, and annotation details are shown in Appendix~\ref{sec:Crowd-sourcing}.}

\paragraph{Analogy Type}
In this paper, we select three types of analogies for generation:
\begin{itemize}[leftmargin=*]
    \item \textbf{Word Analogy}: We adopt the format from \citet{chen-etal-2022-e} in generating word analogies (``A is to B as C is to D'').
    \item \textbf{Structured Analogy}:
    Structured analogies originate from the Structure Mapping Theory~\cite{gentner1997structure}, which posits that analogies are formed by identifying common relational structures between two concepts. 
    Thus, in addition to using one concept to explain another, we also ask the LMs to incorporate related concepts to demonstrate the analogy further.
    \item \textbf{Free-form Analogy}: These analogies utilize unstructured natural language to explain one concept through another. The popularity of this type is increasing with advancements in LLMs~\cite{wijesiriwardene-etal-2023-analogical,ye2024analobench}.
\end{itemize}

Examples of these analogies are provided in Table~\ref{tab:example}, and the prompt templates for generation can be found in Appendix~\ref{sec:generation}.

\subsection{Scientific QA for Student LMs}
In the field of human education, a teacher typically introduces a concept to the class and often uses an analogy to clarify the concept~\cite{thagard1992analogy,heywood2002place,gray2021teaching}. 
For example, when explaining the concept of a cell, drawing an analogy to an automobile factory enhances the understanding, \eg, mitochondria are powerhouses. 
Such analogies help students grasp the concept of a cell, enabling them to correctly answer related questions on homework quizzes.
To align with this, in \method task, given a concept with its analogy generated by teacher LMs, we ask the student LMs to answer questions related to the concept.
The details of the prompt templates are available in Appendix~\ref{sec:answering}.

%

\section{Experiment}

\label{sec:evaluation}

\subsection{Evaluation Protocol}

\paragraph{Evaluation Models}
We choose GPT-4~\cite{openai2023gpt4}, Claude-v3-Sonnet~\cite{anthropic2024claude3}, Mixtral-8x7B~\cite{mistralai_2023_mixtral} as \textbf{teacher LMs}, and GPT-3.5~\cite{openai2022chatgpt}, Gemini~\cite{geminiteam2023gemini}, Mistral-7B~\cite{jiang2023mistral}, Llama3-8B~\cite{llama3modelcard}, Vicuna-13B and Vicuna-7B~\cite{chiang2023vicuna} as \textbf{student LMs}.\footnote{The detailed versions for openai models can be found in Appendix~\ref{sec:model}.}

\paragraph{Evaluation Collection}
We evaluate the models on two datasets that feature various levels of question difficulty:

\begin{itemize}[leftmargin=*]
    \item \textbf{ARC Challenge}~\cite{clark2018think}: This dataset includes 270 natural science questions that stumped both a retrieval-based and a word co-occurrence algorithm.
    \item \textbf{GPQA}~\cite{rein2023gpqa}: With 448 complex multiple-choice questions in biology, physics, and chemistry, it is designed by domain experts. 
    PhD candidates can only achieve 65\% accuracy.
\end{itemize}

\paragraph{Evaluation Metrics}
For all datasets, we report the accuracy of all questions.
Moreover, we randomly sample 100 generated analogies from each dataset and employ three annotators to evaluate their accuracy, with with Fleiss's $\kappa=0.96$~\cite{fleiss1981measurement}.
The annotation details for quality evaluation of generated analogies are shown in Appendix~\ref{sec:Crowd-sourcing}.

\label{sec:analysis}
\subsection{Result \& Analysis}

\begin{table}[t]
\small
  \centering
  \caption{Accuracy (\%) of different student LMs under different strategies. The analogies generated by teacher LMs, \ie, GPT-4, Claude-v3-Sonnet (\textbf{Claude}) and Mixtral-8x7B (\textbf{Mixtral}), are all free-form analogies.}
    \begin{tabular}{lccccc}
    \toprule
    \multirow{2}[0]{*}{\textbf{Student LMs}} & \multirow{2}[0]{*}{\textbf{Direct}} & \multirow{2}[0]{*}{\textbf{CoT}} & \multicolumn{3}{c}{\textbf{Analogy (Teacher LMs)}} \\
    \cmidrule(lr){4-6}
          &       &       & {GPT-4} & {Claude} & {Mixtral} \\
    \midrule
    \rowcolor[gray]{0.95}\multicolumn{6}{c}{\textit{ARC Dataset}}\\
    Gemini & 88.88 & \textbf{89.26} & \textbf{89.26} & 85.56 & 85.18 \\
    GPT-3.5 & 83.33 & 84.44 & \textbf{85.56} & 80.37 & 84.07 \\
    Mistral-7B & 68.52 & 70.74 & \textbf{74.44} & 72.59 & 70.74 \\
    LLama3-8B & 75.19 & 77.04 & 78.89 & \textbf{80.74} & 78.52 \\
    Vicuna-13B & 37.77 & 55.56 & \textbf{63.83} & 61.11 & 62.96 \\
    Vicuna-7B & 25.55 & 34.44 & \textbf{35.56} & 34.44 & 33.42 \\
    
    \midrule
    \rowcolor[gray]{0.95}\multicolumn{6}{c}{\textit{GPQA Dataset}}\\
    Gemini & 41.18  & 41.18  & \textbf{46.41} & 40.52  & 40.52  \\
    GPT-3.5 & 40.32  & 41.83  & \textbf{43.79} & 40.52  & 39.22  \\
    Mistral-7B & 33.33  & 32.68  & \textbf{35.87} & 33.33  & 34.64  \\
    LLama3-8B & 40.52  & 44.44  & \textbf{46.38} & 45.10  & 44.44  \\
    Vicuna-13B & 26.80  & 30.72  & 30.72  & \textbf{32.55} & 30.72  \\
    Vicuna-7B & 24.84  & 18.30  & \textbf{27.45} & \textbf{27.45} & 25.49  \\
    
    \midrule
    \end{tabular}%
  \label{tab:main_result}%
\end{table}%

In the experiments, we expect to answer three research questions:

\paragraph{\textit{RQ1: Can Analogy from Teacher Models Empower Student Models?}}
We adopt Zero-shot Prompting (\textbf{Direct}) and Chain-of-Thought Prompting (\textbf{CoT})~\cite{wei2022chain} as baselines.\footnote{The prompt templates of the two methods are shown in Appendix~\ref{sec:baselines}.}
The results in Table 2 indicate that:
1) Free-form analogies can indeed help student LMs understand scientific concepts better than Zero-shot and CoT Prompting, improving their ability to answer scientific questions. 
2) The analogies generated by GPT-4 improve the ability of student LMs most significantly, indicating the potential of GPT-4 to assist weaker LMs in learning new knowledge.
3) For the GPQA dataset, characterized by specialized concepts and difficult scientific questions, Vicuna-7B and Vicuna-13B perform poorly with Zero-shot and CoT Prompting.
However, with analogies, their performance is effectively enhanced. 
This finding inspires future work to explore using analogies to help the model learn new concepts.

\begin{table}[t]
  \centering
  \small
  \caption{The accuracy (\%) of three types of analogies generated by different teacher LMs. The results are evaluated by human annotators on 100 samples.}
    \begin{tabular}{lcccccc}
    \toprule
    \textbf{Teacher} & \multicolumn{2}{c}{\textbf{Free-form}} & \multicolumn{2}{c}{\textbf{Structured}} & \multicolumn{2}{c}{\textbf{Word}} \\
    \cmidrule(lr){2-3}
    \cmidrule(lr){4-5}
    \cmidrule(lr){6-7}
       \textbf{LMs}   & ARC   & GPQA  & ARC   & GPQA  & ARC   & GPQA \\
    \midrule
    GPT-4 & 100.0  & 94.0  & 100.0  & 98.0  & 100.0  & 100.0  \\
    Claude & 97.0  & 82.0  & 100.0  & 85.0  & 100.0  & 100.0  \\
    Mixtral & 92.0  & 79.0  & 95.0  & 80.0  & 100.0  & 100.0  \\
    \bottomrule
    \end{tabular}%
  \label{tab:analogy_quality}%
\end{table}%

\begin{figure}[t]
    \centering
    \small
    \includegraphics[width=\linewidth]{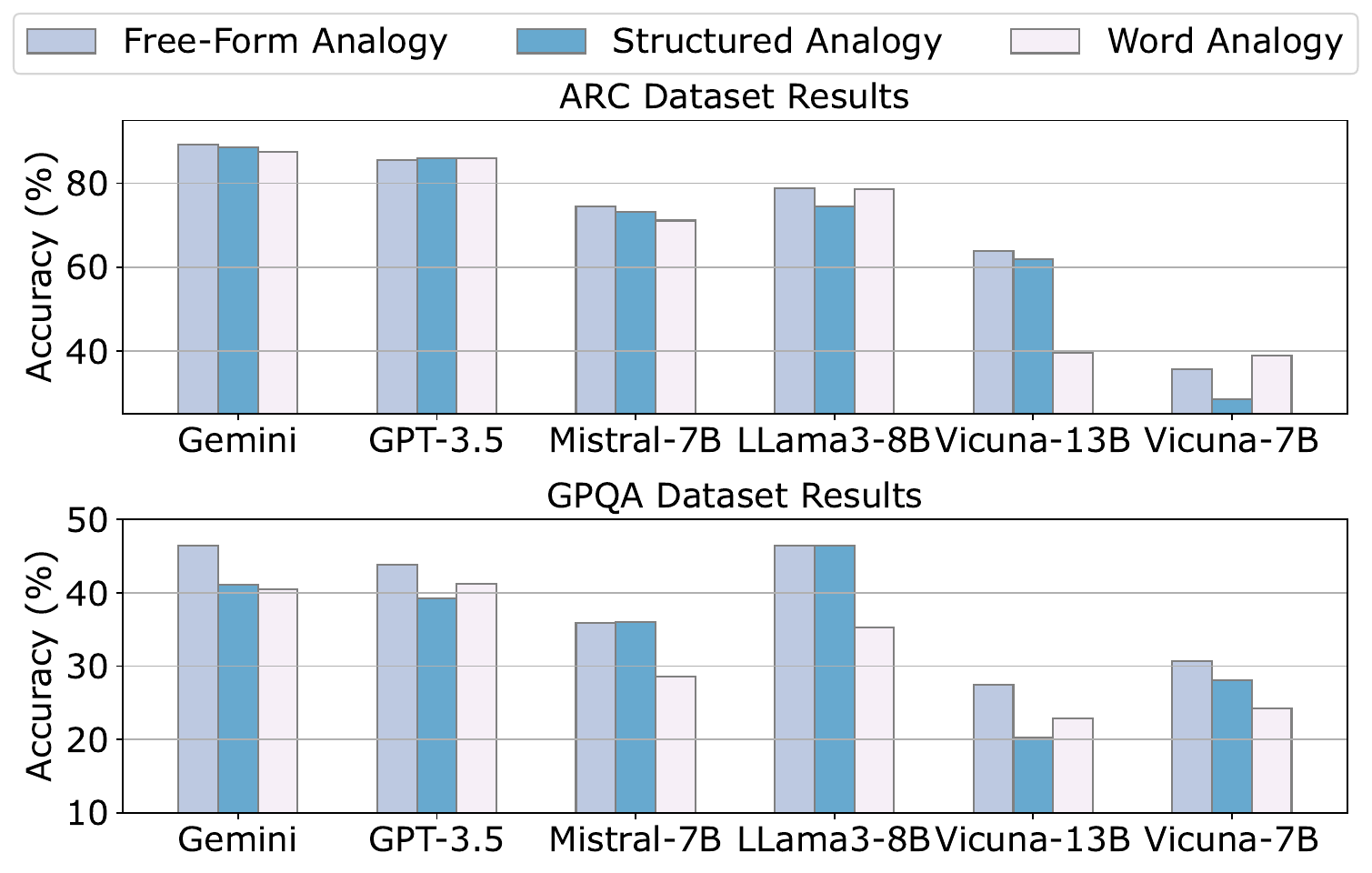}
    \caption{The performance of different student LMs under different types of analogies generated by GPT-4. 
    }
    \label{fig:analogy_type}
\end{figure}

\paragraph{\textit{RQ2: Which Type of Analogy Can Better Empower Student Models?}}

Apart from free-form analogy, we also expect to examine two other analogy types, \ie, structured analogy and word analogy, focusing on their effectiveness in aiding student LMs to grasp scientific concepts. 
As shown in Table~\ref{tab:analogy_quality}, advanced language models such as GPT-4 and Claude-v3-Sonnet and open-source models like Mixtral-8x7B are proficient in generating high-quality word analogies for scientific concepts.
However, the generation quality significantly diminishes for free-form and structured analogies, especially in professional fields (\eg, ``wettability'' and ``contact angle hysteresis'' in the GPQA dataset).

In comparison, Figure~\ref{fig:analogy_type} reveals that compared to word analogy, free-form and structured analogies are more effective in helping models understand scientific concepts due to their more informative content.
Future studies can consider strategies that initially have models generate high-quality word analogies, and then expand them into structured and free-form analogies to enhance their quality.

\begin{table}[t]
  \centering
  \small
  \caption{Comparison of self-generated analogies (\textbf{Analogy$_\texttt{Self}$}) and GPT-4 generated analogies (\textbf{Analogy$_\texttt{GPT-4}$}) for the performance of student LMs on ARC dataset.}
    \begin{tabular}{lcccc}
    \toprule
    \textbf{Model} & \textbf{Direct} & \textbf{CoT} & \textbf{Analogy$_\texttt{Self}$} & \textbf{Analogy$_\texttt{GPT-4}$} \\
    \midrule
    Gemini & 88.88  & 89.26  & 88.88  & \textbf{89.26} \\
    GPT-3.5 & 83.33  & 84.44  & 84.82  & \textbf{85.56} \\
    Mistral-7B & 68.52  & 70.74  & \textbf{77.04 } & 74.44 \\
    LLama3-8B & 75.19  & 77.04  & \textbf{80.37} & 78.89 \\
    Vicuna-13B & 37.77  & 55.56  & 55.93  & \textbf{63.83} \\
    Vicuna-7B & 25.55  & 34.44  & 35.42 & \textbf{35.56} \\
    \bottomrule
    \end{tabular}%
  \label{tab:self_analogy}%
\end{table}%

\paragraph{\textit{RQ3: How About Self-generated Analogy?}}

In addition to using analogies generated by teacher LMs, we also ask student LMs to generate analogies to help themselves understand scientific concepts and answer related questions.
As shown in Table~\ref{tab:self_analogy}, compared to CoT prompting, self-generated analogies can improve the model's understanding of scientific concepts and enhance its ability to answer related questions. 
Moreover, for some models, self-generated analogies outperform those generated by GPT-4, indicating their ability to use analogies to self-learn new knowledge.

\section{Conclusion}
\label{sec:conclusion}

In this paper, we propose the \method task, which simulates the human education process to explore how analogies created by teacher LMs can help student LMs understand scientific concepts. 
Our results suggest that free-form analogies indeed aid LMs in comprehending concepts and enhance their ability to answer related scientific questions accurately. 
Additionally, analogies generated by student LMs can improve their own performance on scientific quizzes, demonstrating their capability to use analogies for self-learning new knowledge.

\section*{Limitations}
\label{sec:limitation}

First, this paper only considers scientific concepts.
We do not cover concepts in other fields, such as historical events and social concepts.
Second, some previous work~\cite{saha2023can} uses explanations generated by stronger LMs to help weaker LMs. 
However, we argue that models may have different strengths in different tasks. 
Therefore, we distinguish between teacher LMs and student LMs without fully evaluating their capabilities. Future work can explore this perspective. 
Additionally, our evaluation is limited to multiple-choice tasks. 
Investigating the performance on more complex tasks, such as RAG, would be beneficial.

\section*{Ethics Statement}
\label{sec:Ethics}
We hereby acknowledge that all authors of this work are aware of the provided EMNLP Code of Ethics and honor the code of conduct.

\paragraph{Use of Human Annotations}
Evaluation on the generated analogies from stronger LMs in \method is implemented by three annotators recruited by our institution.
The construction team remains anonymous to the authors.
We ensure that the privacy rights of all annotators are respected throughout the annotation process. 
All annotators are compensated above the local minimum wage and consent to the use of \method for research purposes, as described in our paper. 
The annotation details are shown in Appendix~\ref{sec:Crowd-sourcing}.

\paragraph{Risks}
The datasets we conduct in the experiment are sourced from publicly available sources, \ie, ARC Challenge Set and GPQA. 
However, we cannot guarantee they are free of socially harmful or toxic language. 
Additionally, analogy evaluation relies on commonsense, and different individuals with diverse backgrounds may have varying perspectives. 
We use ChatGPT to correct grammatical errors in this paper.

\bibliography{anthology,custom}

\clearpage
\appendix

\section{Crowd-sourcing Details}\label{sec:Crowd-sourcing}
\begin{figure*}[t]
    \centering
    \includegraphics[width=\linewidth]{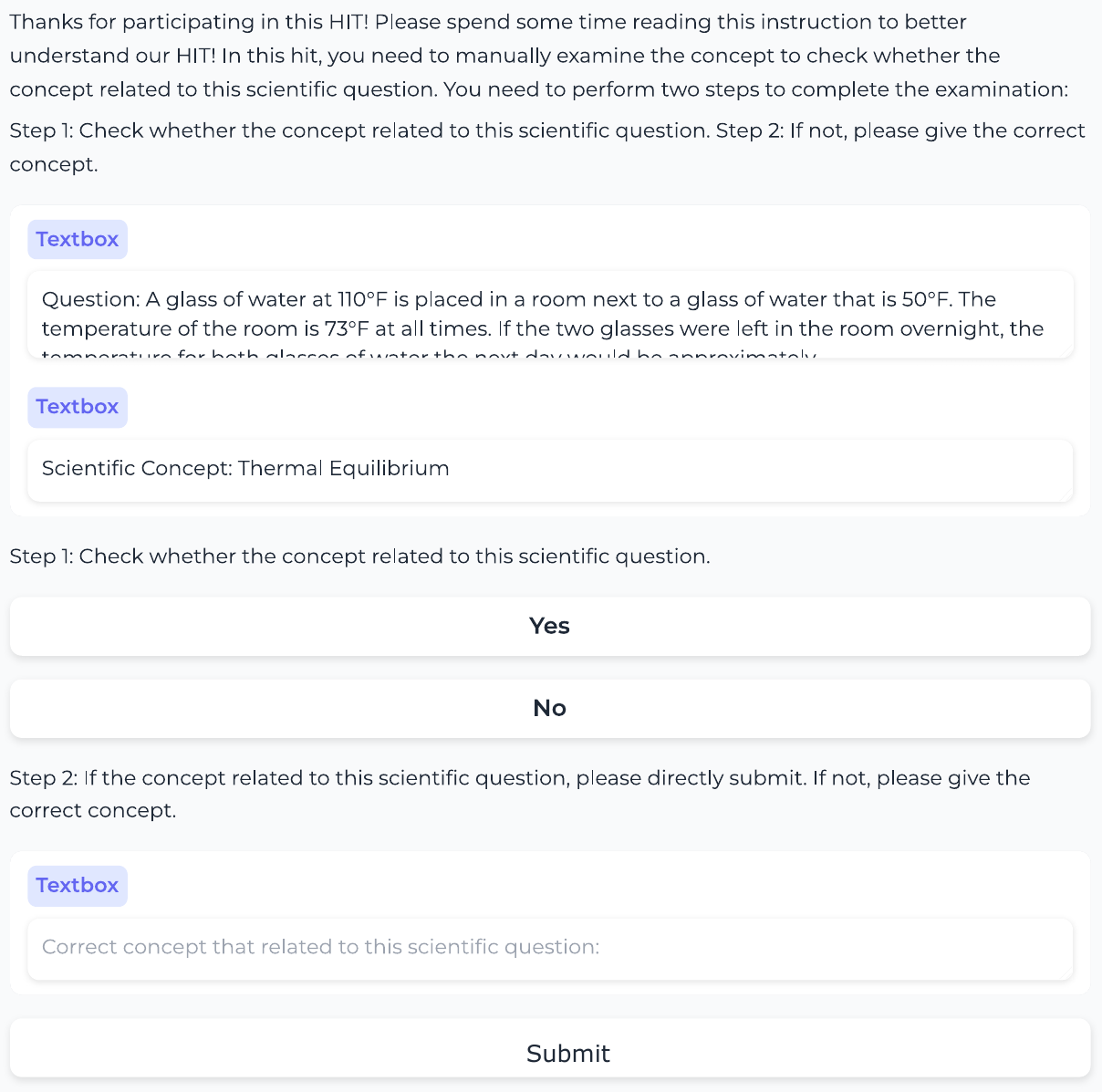}
    \caption{The screenshots of the instructions and interface for extracted concept annotation.}
    \label{fig:concept_annotation}
\end{figure*}

\begin{figure*}[t]
    \centering
    \includegraphics[width=\linewidth]{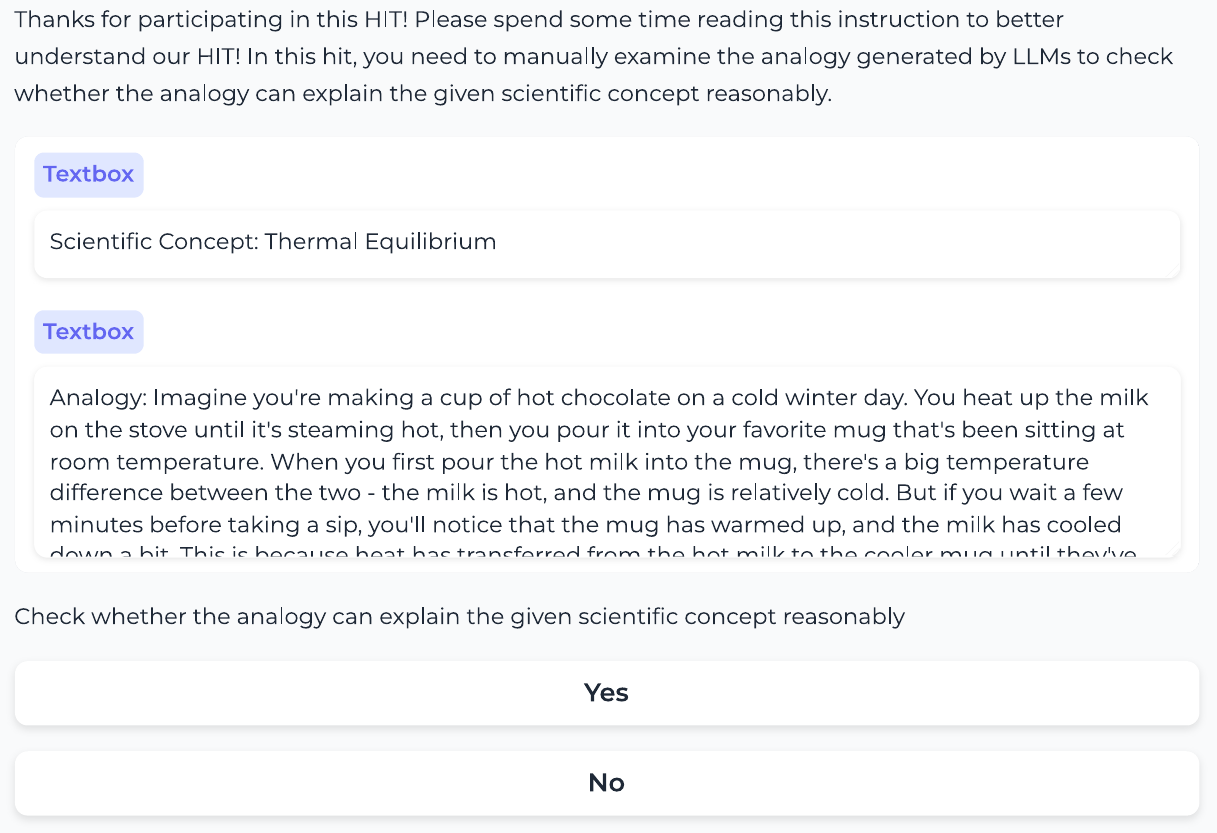}
    \caption{The screenshots of the instructions and interface for generated analogy annotation.}
    \label{fig:analogy_annotation}
\end{figure*}

We have recruited a team of three undergraduates.
To process conflicting annotations, we adopt a voting majority principle to determine the results.
We pay each annotator \$8/h, exceeding the local minimum wage. 
The screenshots of the instructions and interface for quality check of the extracted concepts and generated analogies annotation are shown in Figure~\ref{fig:concept_annotation} and Figure~\ref{fig:analogy_annotation}.

\section{Model Selection}\label{sec:model}
For OpenAI models, we use \texttt{gpt-3.5-turbo-0613} and \texttt{gpt-4-0613}.\footnote{\url{https://platform.openai.com/docs/models}}
For Gemini-pro, we use Google Gemini-Pro APIs to obtain results.
We set the temperature to 0 for all models.

\section{Prompt Template of \method}
\subsection{Analogy Generation}\label{sec:generation}
The prompt of the analogy generation from teacher LMs is given in List~\ref{lst:analogy_generation_prompt}.

\lstset{
    backgroundcolor=\color[RGB]{245,245,244},
    breaklines=true,
    breakindent=0pt,
    basicstyle=\ttfamily\small,
    emph={Structure,Analogy,Free,Form,Word,Generation,Concept},
    emphstyle={\bfseries\color{NavyBlue}}
}\begin{lstlisting}[caption={Instruction templates for teacher LMs to generate analogies.},label=lst:analogy_generation_prompt]
Free-Form Analogy Generation:
Please use an analogy with no more than 100 words to explain the scientific concept:
Example:
Concept: Thermal Equilibrium
Analogy: Imagine you're making a cup of hot chocolate on a cold winter day. You heat up the milk on the stove until it's steaming hot, then you pour it into your favorite mug that's been sitting at room temperature. When you first pour the hot milk into the mug, there's a big temperature difference between the two - the milk is hot, and the mug is relatively cold. But if you wait a few minutes before taking a sip, you'll notice that the mug has warmed up, and the milk has cooled down a bit. This is because heat has transferred from the hot milk to the cooler mug until they've reached a point where they're the same temperature. This is thermal equilibrium.Just like the hot milk and the mug, when two objects at different temperatures come into contact, heat will always flow from the hotter object to the cooler one. This continues until they reach thermal equilibrium, or the same temperature. Once they're at the same temperature, there's no more heat flow because there's no temperature difference to drive it.
Concept: {scientific_concept}
Analogy:

Structure Analogy Generation:
Given one scientific concept, you should use another concept as an analogy to explain this concept. Moreover, you should use other concepts that are related to these two concept to explain the analogy:
Example:
Concept: Thermal Equilibrium
Analogy: Thermal Equilibrium can be analogous to a Balancing Scale
1. Hot and cold objects correspond to weights on a scale: Just as a hot object and a cold object interact to reach thermal equilibrium, weights on a scale interact to reach a balanced state. The hot object, like a heavier weight, has an excess (of heat or weight) that it transfers to the cold object or lighter weight.
2. Heat transfer corresponds to weight redistribution: In thermal equilibrium, heat transfers from the hot object to the cold object until they reach the same temperature. Similarly, on a balancing scale, weight redistributes from the heavier side to the lighter side until they reach the same level. 
3. The point of equilibrium corresponds to the balance point on a scale: In thermal equilibrium, the point of equilibrium is when both objects reach the same temperature. On a balancing scale, the balance point is reached when both sides of the scale are at the same level, indicating that the weights are equal.
4. The cessation of heat flow corresponds to the stillness of the scale: Once thermal equilibrium is reached, there is no more heat flow because there's no temperature difference to drive it. Similarly, once a scale is balanced, there is no more movement because there's no weight difference to drive it.
Concept: {scientific_concept}
Analogy:


Word Analogy Generation:
Given one scientific concept, you should use another concept as an analogy to explain this concept:
Example:
Concept: Thermal Equilibrium
Analogy: Thermal Equilibrium can be analogous to a Balancing Scale
Concept: {scientific_concept}
Analogy:
\end{lstlisting}

\subsection{Question Answering}\label{sec:answering}
The prompt of the question answering by student LMs is given in List~\ref{lst:question_answering}.

\lstset{
    backgroundcolor=\color[RGB]{245,245,244},
    breaklines=true,
    breakindent=0pt,
    basicstyle=\ttfamily\small,
    emph={Structure,Analogy,Free,Form,Word,Generation,Concept},
    emphstyle={\bfseries\color{NavyBlue}}
}\begin{lstlisting}[caption={Instruction templates for student LMs to answer questions based on analogies.},label=lst:question_answering]
You need to select an answer for a question.
This is the question: 
{question}
{choices}
Since the question is difficult, we asked a teacher to explain the concepts in this question to you using analogies, which we hope can help you.
This is the explanation with analogies:
{analogy}
Please combine the explanation to better answer this question.
Answer:   
\end{lstlisting}

\subsection{Concept Extraction}\label{sec:extraction}
The prompt of the concept extraction by GPT-4 is given in List~\ref{lst:concept_extraction}.

\lstset{
    backgroundcolor=\color[RGB]{245,245,244},
    breaklines=true,
    breakindent=0pt,
    basicstyle=\ttfamily\small,
    emph={Structure,Analogy,Free,Form,Word,Generation,Concept},
    emphstyle={\bfseries\color{NavyBlue}}
}\begin{lstlisting}[caption={Instruction templates for GPT-4 to extract scientific concepts.},label=lst:concept_extraction]
Given a scientific question, you should show the key scientific concept related to this scientific question.
This is a scientific question:
{question}
The key scientific concept:
\end{lstlisting}

\subsection{The Prompt Templates of Zero-shot and CoT Prompting}\label{sec:baselines}
The prompt of Zero-shot and CoT Prompting is given in List~\ref{lst:baselines}.

\lstset{
    backgroundcolor=\color[RGB]{245,245,244},
    breaklines=true,
    breakindent=0pt,
    basicstyle=\ttfamily\small,
    emph={Zero,shot,Prompting,CoT},
    emphstyle={\bfseries\color{NavyBlue}}
}\begin{lstlisting}[caption={Instruction templates for Zero-shot and CoT Prompting.},label=lst:baselines]
Zero-shot Prompting:
{question}
{Options}
Answer:

CoT Prompting:
{question}
{Options}
You need to give the reason first and then choose the answer.
Answer:
\end{lstlisting}

\label{sec:appendix}
\end{document}